\documentclass[letterpaper, 10 pt, conference]{ieeeconf}  % Comment this line out if you need a4paper

\IEEEoverridecommandlockouts                              % This command is only needed if 
\usepackage{dsfont}
\usepackage{amsmath}
\usepackage{graphicx}
\usepackage{xcolor}
\usepackage{algorithm}
\usepackage{algpseudocode}
\usepackage[compress]{cite}

\newcommand{\ProcessModel}{\mathcal{P}}
\newcommand{\MeasurementModel}{\mathcal{M}}
\newcommand{\Dynamics}{f}
\newcommand{\Cost}{J}
\newcommand{\Constraint}{C}

\newcommand{\GameSolver}{\mathcal{G}}
\DeclareMathOperator*{\argmin}{argmin}

\title{\LARGE \bf
AToM: Adaptive Theory-of-Mind-Based Human Motion Prediction in Long-Term Human-Robot Interactions
}

\author{Yuwen Liao$^{1}$, Muqing Cao$^{1}$, Xinhang Xu$^{1}$, Lihua Xie$^{1}$, $\textsl{Fellow}$, $\textsl{IEEE}$% <-this % stops a space
\thanks{$^{1}$Authors are with the School of Electrical and Electronic Engineering, 
        Nanyang Technological University, Singapore.
        {\tt\small \{yuwen001, mqcao, xu0021ng, elhxie\}@ntu.edu.sg}}%
\thanks{This research is partially supported by the Centre for Advanced Robotics Technology Innovation (CARTIN) funded by the National Research Foundation of Singapore under its Medium-Sized Centre Scheme, and the Agency for Science, Technology and Research (A*STAR) of Singapore under the grant M21K1a0104.}%
}

\begin{document}

\maketitle
\thispagestyle{empty}
\pagestyle{empty}

%%%%%%%%%%%%%%%%%%%%%%%%%%%%%%%%%%%%%%%%%%%%%%%%%%%%%%%%%%%%%%%%%%%%%%%%%%%%%%%%
\begin{abstract}

Humans learn from observations and experiences to adjust their behaviours towards better performance. 
Interacting with such dynamic humans is challenging, as the robot needs to predict the humans accurately for safe and efficient operations. 
Long-term interactions with dynamic humans have not been extensively studied by prior works. 
We propose an adaptive human prediction model based on the Theory-of-Mind (ToM), a fundamental social-cognitive ability that enables humans to infer others' behaviours and intentions. 
We formulate the human internal belief about others using a game-theoretic model, which predicts the future motions of all agents in a navigation scenario. 
To estimate an evolving belief, we use an Unscented Kalman Filter to update the behavioural parameters in the human internal model.
% To correct and develop this estimated belief, we use an Unscented Kalman Filter to update the behavioural parameters in the human internal model.
% The predicted human trajectories can be used in robot predictive planning, and the human-predicted robot trajectories reflect how the human infer the robot behaviours using ToM.
% We show that our model can generate accurate human predictions and provide unique interpretability to dynamic human behaviours. 
Our formulation provides unique interpretability to dynamic human behaviours by inferring how the human predicts the robot.
We demonstrate through long-term experiments in both simulations and real-world settings that our prediction effectively promotes safety and efficiency in downstream robot planning. Code will be available at https://github.com/centiLinda/AToM-human-prediction.git.

\end{abstract}

\section{Introduction}

Autonomous robots are frequently deployed in smart factories and workplaces to accomplish tasks with human workers. An important aspect of real-life human-robot interactions is that human behaviours do not stay identical during long-term interactions (i.e., repeated interactions under the same scenario \cite{sagheb2023towards}). 
Consider an example where a robot and a human move in a shared space. During the initial encounters, the human might keep a large distance away from the robot due to a lack of understanding of its behaviours. 
%After working together for some time, the human observes that the robot is programmed with collision-avoidance capability and starts to move more efficiently with confidence that the robot is safe to interact with. 
%Since the human now accepts a smaller social distance, the robot plan can become more efficient if the robot can capture this dynamic human behaviour. The robot movement will remain inefficient if it sticks to the initial human prediction.
After observing the robot's consistent collision-avoidance capabilities, the human's confidence in its safety grows, leading to closer and more efficient interactions. 
Capturing this shift in human perception allows the robot to better predict human motion, enabling safer and more efficient navigation.

Existing human trajectory predictors are incapable of modelling such dynamic human behaviours. Rule-based human predictors are defined using heuristics and common senses, 
which often produce unrealistic human trajectories and may fail to predict complex motions \cite{rudenko2020human, korbmacher2022review}. Data-driven neural networks are trained on large-scale datasets to model common pedestrian behaviours \cite{sighencea2021review}. 
These methods can be fine-tuned for a specific scenario or dataset, but do not adaptively predict human trajectories in long-term human-robot interactions.

% Autonomous robots are frequently deployed in smart factories and workplaces to accomplish tasks with human workers. During such collaborations, the robots need to accurately predict human motions in order to navigate in a shared space safely and efficiently. An important aspect of real-life human-robot interactions is that human behaviours do not stay identical during long-term and repeated interactions. Most existing human prediction methods cannot model these dynamic human behaviours. Some approaches attempt to quantify the uncertainty in human motions by providing probabilistic predictions, but such predictions might cause the robot to be over-conservative, and they cannot explain why humans change their behaviours.
% {\color{blue}You explained about long-term human interaction but it is a bit abstract. Need to explain in the context of motion planning why considering long-term interaction is important. Give example of how human's motion changes in repeated encounters. Mention that the change in the behaviour may be modelled by certain parameters such as velocity and distance.}

\begin{figure}
    \centering
    \includegraphics[width=\linewidth]{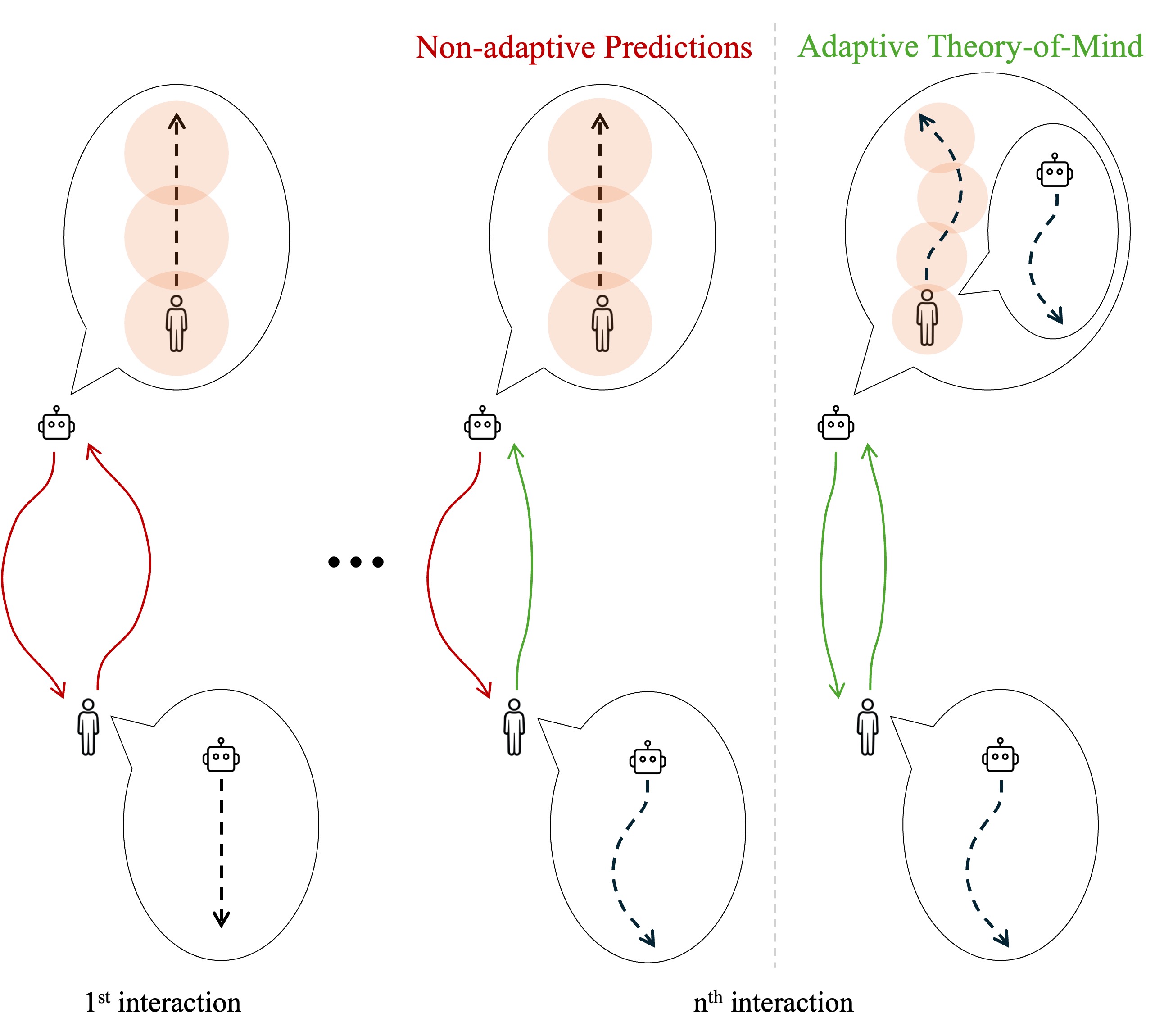}
    \vspace{-7mm}
    \caption{After repeated interactions, the human develops confidence to move in a closer and more efficient manner. The robot, on the other hand, remains inefficient using a non-adaptive human model. Our proposed adaptive Theory-of-Mind (AToM) captures evolving human internal beliefs of others, allowing the robot to plan a more efficient path.}
    \vspace{-7mm}
    \label{fig:illustration}
\end{figure}

We take inspiration from human-human interactions and adopt a term from psychology called Theory-of-Mind (ToM). It states that humans have the mental capability to infer the emotions, thoughts, and intentions of others. 
Through observations and interactions, humans can update and correct their understanding of others' behaviours, therefore modify their own decisions in future interactions. 
As illustrated in Fig. \ref{fig:illustration}, the human learns to move in a more efficient way by developing a belief that the robot is capable of collision avoidance.

We thus seek to predict dynamic human motions in human-robot interactions using the ToM model. Existing works have used ToM to model the decision-making process in extensive-form games such as rock-paper-scissors \cite{de2013much}. 
However, to predict human trajectories for robot planning, directly applying ToM results in a recursive structure (the robot plans from human prediction, which in turn depends on the robot plan) that is difficult to solve. 
We propose an innovative reformulation of the original ToM. We use a game-theoretic model as the human internal belief model. 
Both the human and robot trajectories are controlled by a set of behavioural parameters such as speed limits and preferred safe distances. 
By solving for a Nash equilibrium solution, we obtain the human-predicted robot trajectory and predicted human trajectory, the latter can be used in downstream robot predictive planning. 
To correct this estimated human internal model, we update the behavioural parameters using an Unscented Kalman Filter. 
Our adaptive ToM model has no access to either the robot planner or the true human model (which does not exist), but it can improve prediction accuracy and adjust to dynamic human behaviours through the update process. 
Our model also provides a unique insight by explicitly predicting how humans infer the behaviours of surrounding agents.

Of relevance to our work is the literature on interactive and dynamic human prediction. 
The former considers the influence of robots on humans by either treating the robot as a human neighbour, or treating the human as a follower that reacts to the robot leader in a Stackelberg game. 
The latter corrects noisy parameters in a human model to approach some true preset values. 
Our work is fundamentally different from these methods as 1) we enable the human to ``predict" apart from ``observe", 2) we do not assume that there exists a true static human model and instead estimate a dynamic human.

Our contributions are threefold. 
First, we propose a ToM-based human predictor that models human internal beliefs about surrounding agents. 
Second, we incorporate an update mechanism to capture dynamic human behaviours in long-term human-robot interactions. 
Third, we demonstrate through comparisons that our method provides more accurate trajectory predictions and promotes safety and efficiency in downstream robot planning.

\section{Related Work}
% We first introduce some existing works about modelling an agent's decision-making process using ToM. We then review two groups of methods that focus on modelling interactive and dynamic humans, respectively. Lastly, we review common types of human trajectory prediction methods used in robot planning.

\subsection{ToM in Agent Modelling}
\label{sec:ToM}
% ToM has been a well-established theory in cognitive psychology \cite{baron1997mindblindness}. 
Prior works have formulated a ToM agent's decision-making processes for basic games such as rock-paper-scissors \cite{de2013much}, Tacit Communication Game \cite{de2015higher}, and Common Pool Resource Game \cite{von2017minds}. 
These extensive-form games have relatively small action sets, making recursive reasoning straightforward by listing down all possible game states. Some methods apply ToM to predict agent's movement \cite{baker2014modeling, rabinowitz2018machine, tian2021learning}, but they are constrained to a grid space with limited action sets. For a more general discussion on ToM in machine intelligence, we direct the reader to \cite{cuzzolin2020knowing}.

\subsection{Interactive Human Models}
Many human motion predictors have studied interactions in human crowds \cite{sighencea2021review}. A heuristic approach for human-robot interaction is to directly model the robot as a human neighbour. 
\cite{schaefer2021leveraging} uses a neural network to predict human trajectories with different candidate robot plans. By observing how much the human tries to avoid each planned trajectory, it chooses the least invasive robot plan for maximised human comfort. 
This method relies on a strong assumption that humans treat robots the same as other humans. 
Another line of work models how humans react to robots. \cite{tian2022safety} predicts whether the human behaves like a leader or follower in a Stackelberg game. When the human is a follower, the cost is minimised based on observed robot actions. 
Similarly in \cite{geldenbott2024legible, sripathy2021dynamically}, the robot action is used in human cost functions to predict the human response. The key difference between these methods and our proposed method is that we enable the human to ``predict" future motions of other agents, instead of only ``observe and respond".

\subsection{Dynamic Human Models}
Instead of using a unified and static human model for all interactions, some methods model a dynamic human. 
\cite{tian2023towards} models a human teleoperator whose internal estimation of the robot dynamics is potentially inaccurate. The robot then corrects this estimation by modifying its response to teleoperation. It assumes that there exists a true value that the human model needs to approach. However, there is no true human model in reality. 
\cite{parekh2023learning, muktadir2024adaptive, cathcart2023proactive} models diverse human actions and preferences using either explicit categories or latent distributions.
% \cite{parekh2023learning} models different human behaviours using a latent distribution. Future human actions are predicted from the most likely latent behaviour sample. 
% \cite{muktadir2024adaptive} quantitatively defines human driver behaviours to categorize observed vehicle motions. Future motions are predicted separately for each behaviour group. 
These methods provide some level of interpretability of different human behaviours, but they fail to explain why humans change behaviours over repeated interactions.

\subsection{Human Trajectory Prediction for Robot Planning}
\label{seq:traj_prediction}
% To navigate safely in a shared space with humans, robots rely on human trajectory predictions to perform predictive planning. 
Early works on robot planning do not consider human motion models \cite{wu2019depth, wu2018learn, wu2021learn, wu2020achieving, wu2019bnd, wu2019tdpp, cao2022direct}.
In recent social predictive planning, the most commonly adopted heuristic is the constant velocity model. 
Another classic model is the Social Force model \cite{helbing1995social} which considers multiple factors that can influence human motions. 
Recent learning-based methods have achieved outstanding prediction accuracy on large-scale pedestrian datasets \cite{salzmann2020trajectron++, xu2022remember}. 
Existing robot planners have integrated these human predictors into the system \cite{cao2024learningdynamicweightadjustment, boldrer2022multi, boldrer2020socially, ryu2024integrating, poddar2023crowd}. 
However, few methods have compared the effect of different prediction methods over downstream planning. 
Through our experiment, we show that our model not only produces more accurate predictions, but also ensures safety and efficiency in robot planning.

\begin{figure*}[ht]
    \centering
    \includegraphics[width=0.95\textwidth]{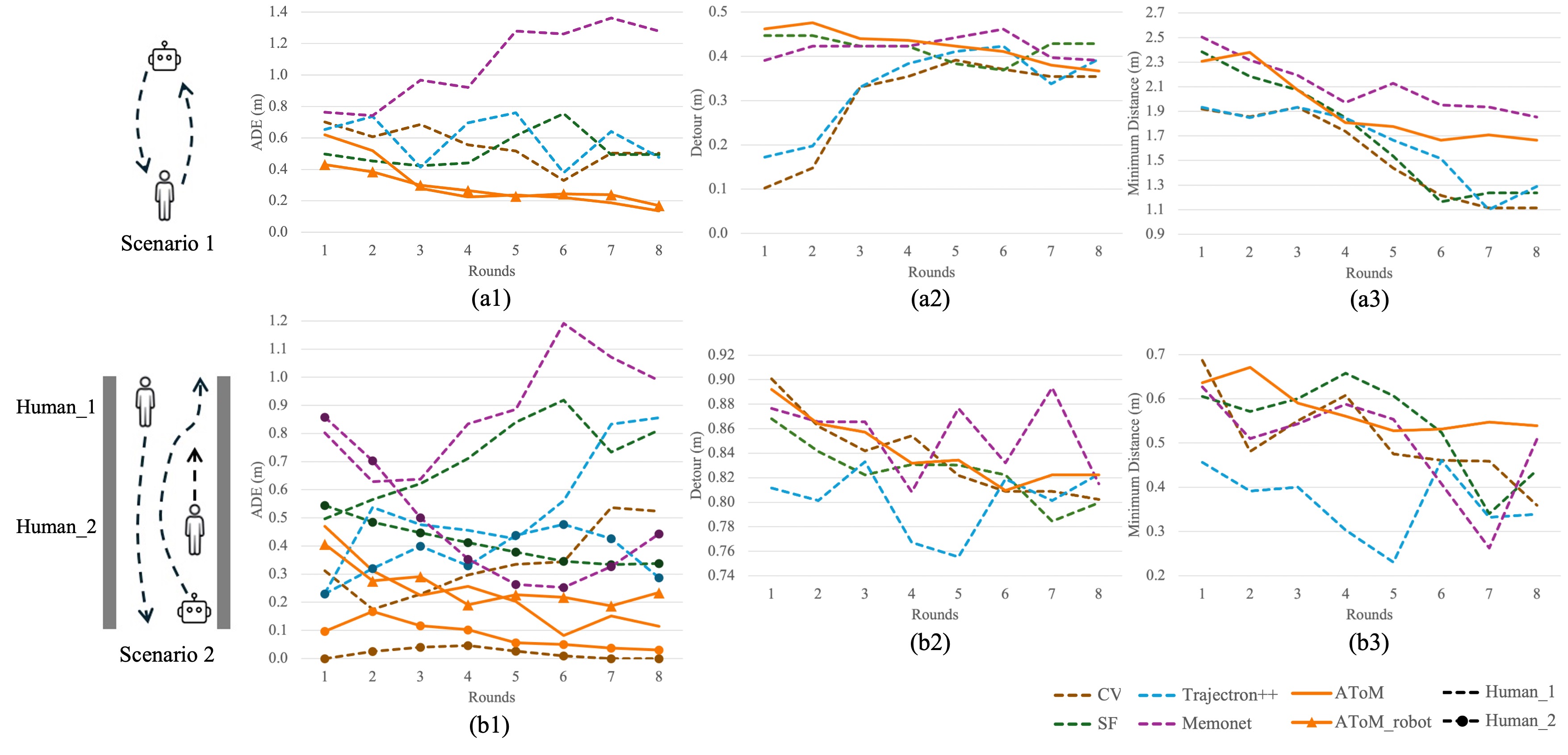}
    \vspace{-5mm}
    \caption{Quantitative comparisons for Scenario 1 and Scenario 2. The simulation setup is illustrated on the left. We compare the prediction accuracy using ADE, and the efficiency and safety in resulting robot plans using Detour and Minimum Distance.}
    % AToM achieves the best accuracy and results in improving efficiency while maintaining safety.}
    \vspace{-5mm}
    \label{fig:quantitative_12}
\end{figure*}

\section{Methodology}
\textbf{Problem Definition.} 
Let the state and control of an agent be $\boldsymbol{x} \in \mathds{R}^{n}$ and $\boldsymbol{u} \in \mathds{R}^{m}$, where $n, m$ are state and control dimensions. 
We use subscripts $H, R, HR$ to represent human, robot, and their joint system, and superscript $k$ to represent timesteps.
The dynamics can be described by $\boldsymbol{x}^{k+1} = \Dynamics(\boldsymbol{x}^{k}, \boldsymbol{u}^{k})$.
The trajectories from timestep $i$ to timestep $j$ can be represented by $\boldsymbol{X}^{i:j} = [\boldsymbol{x}^{i}, \dots, \boldsymbol{x}^{j}]$.
In a long-term interaction setting, we assume that the goal positions and environmental obstacles are known as $\boldsymbol{g}_{HR}$ and $\mathcal{O}$, respectively.
At each timestep $t$, given the observed historical trajectories $\boldsymbol{X}_{HR}^{0:t}$ and known information $\boldsymbol{g}_{HR}$ and $\mathcal{O}$, our goal is to predict future human trajectories $\hat{\boldsymbol{X}}_{H}^{t+1:t+T_{f}}$ with prediction horizon $T_{f}$, which the robot can use for collision-free predictive planning. 
For simplicity, we will omit time superscripts when no confusion is aroused. Our method is planner-agnostic and the robot planning algorithm is outside the scope of this work. 

\textbf{Overview.} 
We first present the original nested ToM formulation in robot predictive planning. We then propose our reformulated adaptive ToM (AToM) which detaches the human prediction model from the recursive structure. The human model consists of a game-theoretic solver and a parameter update mechanism, which captures dynamic human behaviours in long-term interactions.

\subsection{Original ToM in Robot Planning}
Following prior works mentioned in Sec. \ref{sec:ToM}, we formulate the original ToM in robot predictive planning task and identify the potential issues with this formulation.

The robot plans with a finite horizon $T_{p}$, where at each timestep t the robot optimises its controls $\boldsymbol{U}_{R} = [\boldsymbol{u}_{R}^{t+1}, \dots, \boldsymbol{u}_{R}^{t+T_{p}}]$ to minimize the cumulative cost:
\begin{equation}
\label{eq:robot_planning}
\begin{aligned}
    \min &\; \Cost_{R}(\boldsymbol{x}_{R}^{t+1:t+T_{p}}, \boldsymbol{x}_{R}^{t}, \hat{\boldsymbol{X}}_{H}, \boldsymbol{g}_{R}, \mathcal{O}), \\
    \text{s.t.} \quad &\boldsymbol{x}_{R}^{k+1} = \Dynamics_{R}(\boldsymbol{x}_{R}^{k}, \boldsymbol{u}_{R}^{k}), \\
                      &\Constraint_{R}(\boldsymbol{x}_{R}^{k}, \boldsymbol{u}_{R}^{k}, \boldsymbol{x}_{H}^{k}, \mathcal{O}) \leq 0, \\
                      &k = t+1, \dots, t+T_{p},
\end{aligned}
\end{equation}
where $\Cost_{R}$ is the cost function that captures the robot's performance goals, 
$\Constraint_{R}$ contains constraints such as collision-avoidance, 
and predicted human trajectories $\hat{\boldsymbol{X}}_{H}$ can be obtained from human prediction models such as the Social Force or prediction neural networks, as described in Sec. \ref{seq:traj_prediction}.

% Human predictions $\hat{X}_{H}$ can be obtained from historical observations and other additional information such as human goal and environmental obstacles: ({\color{blue} can introduce the parameter here and remove equation \eqref{eq:humanpred2}})
% \begin{equation}
% \label{eq:human_pred}
%     \tilde{X}_{H} = f_{pred} (X_{HR}, g_{H}, \mathcal{O}),
% \end{equation}
% where $f_{pred}$ is the prediction method such as the Social Force model or neural networks. 

To obtain $\hat{\boldsymbol{X}}_{H}$ using a ToM model, predicted human actions need to be solved from predicted robot trajectories:
\begin{equation}
\label{eq:original_ToM}
\begin{aligned}
    \boldsymbol{\hat{U}}_{H} = \argmin_{\boldsymbol{u}_H^{t+1:t+T_{f}}} &\Cost_{H}(\boldsymbol{x}_{H}^{t+1:t+T_{f}}, \boldsymbol{x}_{HR}^{t}, \hat{\boldsymbol{X}}_{R}, \boldsymbol{g}_{H}, \mathcal{O}), \\
    \text{s.t.} \quad &\boldsymbol{x}_{H}^{k+1} = \Dynamics_{H}(\boldsymbol{x}_{H}^{k}, \boldsymbol{u}_{H}^{k}), \\
                      &\boldsymbol{\hat{x}}_{R}^{k+1} = \Dynamics_{R}(\boldsymbol{\hat{x}}_{R}^{k}, \boldsymbol{u}_{R}^{k}), \\
                      &\Constraint_{H}(\boldsymbol{x}_{H}^{k}, \boldsymbol{u}_{H}^{k}, \boldsymbol{x}_{R}^{k}, \mathcal{O}) \leq 0, \\
                      &k = t+1, \dots, t+T_{f}. \\
\end{aligned}
\end{equation}
% In the original ToM formulation, future human actions are optimised similarly to robot predictive planning:
% \begin{equation}
% \label{eq:original_ToM}
% \begin{aligned}
%     \min_{u_H^{t+1:t+T_{pred}}} &\sum_{i=t+1}^{t+T_{pred}} J_{H}(x_{H}^{i}, g_{H}, \tilde{X}_{R}, \mathcal{O}), \\
%     \text{s.t.} \quad &x_{H}^{i+1} = x_{H}^{i} + f_{H}(x_{H}^{i}, u_{H}^{i}), \\
%                       &x_{R}^{i+1} = x_{R}^{i} + f_{R}(x_{R}^{i}, u_{R}^{i}), \\
%                       &C(x_{H}^{i}, u_{H}^{i}, x_{R}^{i}) \leq 0, \\
%                       &i = t+1, \dots, t+T_{pred},
% \end{aligned}
% \end{equation}
Note that the predicted robot trajectories are obtained from the robot plans $\boldsymbol{U}_{R}$ solved from Problem \eqref{eq:robot_planning}. 
If we substitute Eq. \eqref{eq:original_ToM} and human dynamics $\Dynamics_{H}$ back to the cost in Problem \eqref{eq:robot_planning}, it results in a recursive optimisation problem for the robot. 
Solving such optimisation is tractable for simple settings as discussed in Sec. \ref{sec:ToM}, but complex and computationally expensive for robot planning with continuous and infinite action space. 
Therefore, we propose a reformulation of the original ToM to detach the human prediction model from the nested optimisation problem.

% \begin{figure}
%     \centering
%     \includegraphics[width=\linewidth]{quantitative_3.jpg}
%     \vspace{-8mm}
%     \caption{Quantitative comparison between AToM and SF in Scenario 3. We compare the number of steps the robot takes to reach the goal which reflects the efficiency. We mark the rounds where collisions happens using red crosses.}
%     % AToM enables the robot to pass through the doorway safely and efficiently in all rounds of experiments.}
%     \vspace{-5mm}
%     \label{fig:quantitative_3}
% \end{figure}

\subsection{Reformulation and Adaptive ToM (AToM)}
Instead of using the true robot plan as human-predicted robot actions, we argue that the human maintains an internal model of the navigation problem and they predict the optimal actions for all agents based on a dynamic belief about each agent's behavioural pattern. 
We formulate the human internal model as a multi-player general-sum differential game with finite horizon $T_{f}$. 
In a navigation scenario, the strategy for player $i$ is a control sequence $\boldsymbol{U}_{i} = [\boldsymbol{u}_{i}^{t+1}, \dots, \boldsymbol{u}_{i}^{t+T_{f}}]$. 
Each player seeks to minimize its cost while respecting constraints and dynamics:
\begin{equation}
\label{eq:game_cost}
\begin{aligned}
    \min_{U} \quad &\sum_{k=t+1}^{t+T_{f}} 
    (\boldsymbol{x}_{i}^{k} - \boldsymbol{g})^\top Q (\boldsymbol{x}_{i}^{k} - \boldsymbol{g}) 
    + \boldsymbol{u}_{i}^{k}{}^\top R \boldsymbol{u}_{i}^{k} \\
    +  &w_{s}(\sum_{n \neq i} \max (0, \|\boldsymbol{x}^{k}_{i} - \boldsymbol{x}^{k}_{n}\|_{2} - d_{s}))) \\
    + &w_{o}(\max (0, D(\boldsymbol{x}^{k}_{i}, \mathcal{O}) - d_{o}))\\
    \text{s.t.} \quad &\boldsymbol{x}_{min} \leq \boldsymbol{x}_{i}^{k} \leq \boldsymbol{x}_{max}, \\
                &\boldsymbol{u}_{min} \leq \boldsymbol{u}_{i}^{k} \leq \boldsymbol{u}_{max}, \\
                &\boldsymbol{x}_{i}^{k+1} = \Dynamics_{i}(\boldsymbol{x}_{i}^{k}, \boldsymbol{u}_{i}^{k}), \\
                &k = t+1, \dots, t+T_{f}, \\
\end{aligned}
\end{equation}
% \begin{equation}
% \label{eq:game_cost}
% \begin{aligned}
%     \min_{U} \quad & \Cost_{goal}(\boldsymbol{X}, \boldsymbol{g}, \theta_{1}) + \Cost_{speed}(\boldsymbol{X}, \theta_{2}) \\
%                    &+ \Cost_{obs}(\boldsymbol{X}, \mathcal{O}, \theta_{3}) + \Cost_{social}(\boldsymbol{X}_{joint}, \theta_{4}), \\
%     \text{s.t.} \quad &\Constraint(\boldsymbol{X}_{joint}, \boldsymbol{U}, \mathcal{O}, \theta_{5}) \leq 0, \\
%                 &\boldsymbol{x}^{k+1} = \Dynamics(\boldsymbol{x}^{k}, \boldsymbol{u}^{k}), \\
%                 &k = t+1, \dots, t+T_{f}, \\
% \end{aligned}
% \end{equation}
% where the 4 cost components are $\Cost_{goal}$ for goal reaching, $\Cost_{speed}$ for speed regulation, $\Cost_{obs}$ for obstacle avoidance, and $\Cost_{social}$ for social distancing. 
% $\boldsymbol{\theta} = [\theta_{1}, \theta_{2}, \theta_{3}, \theta_{4}, \theta_{5}]^\top$ are the parameters from the costs and constraints 
% such as the preferred speed in $\Cost_{speed}$, and preferred distance from neighbours in $\Cost_{social}$. 
where $Q \geq 0$ and $R > 0$ are the weights for the state and control, $w_{s}$ and $w_{o}$ are the weights for social and obstacle avoidance, $D$ calculates the closest distance to obstacle $\mathcal{O}$, 
$d_{s}$ and $d_{o}$ are the preferred social and obstacle distances, $\boldsymbol{x}_{min}$, $\boldsymbol{x}_{max}$, $\boldsymbol{u}_{min}$, and $\boldsymbol{u}_{max}$ are the state and control limits.
We represent these behavioural parameters for all players using $\boldsymbol{\theta}$.
% $\boldsymbol{\theta} = [w_{s}, d, w_{o}, d_{o}, \boldsymbol{x}_{min}, \boldsymbol{x}_{max}, \boldsymbol{u}_{min}, \boldsymbol{u}_{max}]^\top$ are the parameters from the costs and constraints such as the weights, preferred distance from other agents and obstacles, and limits for the states and controls.
% These parameters model the behaviours of each player in a game-theoretic setting.

\begin{figure}
    \centering
    \includegraphics[width=\linewidth]{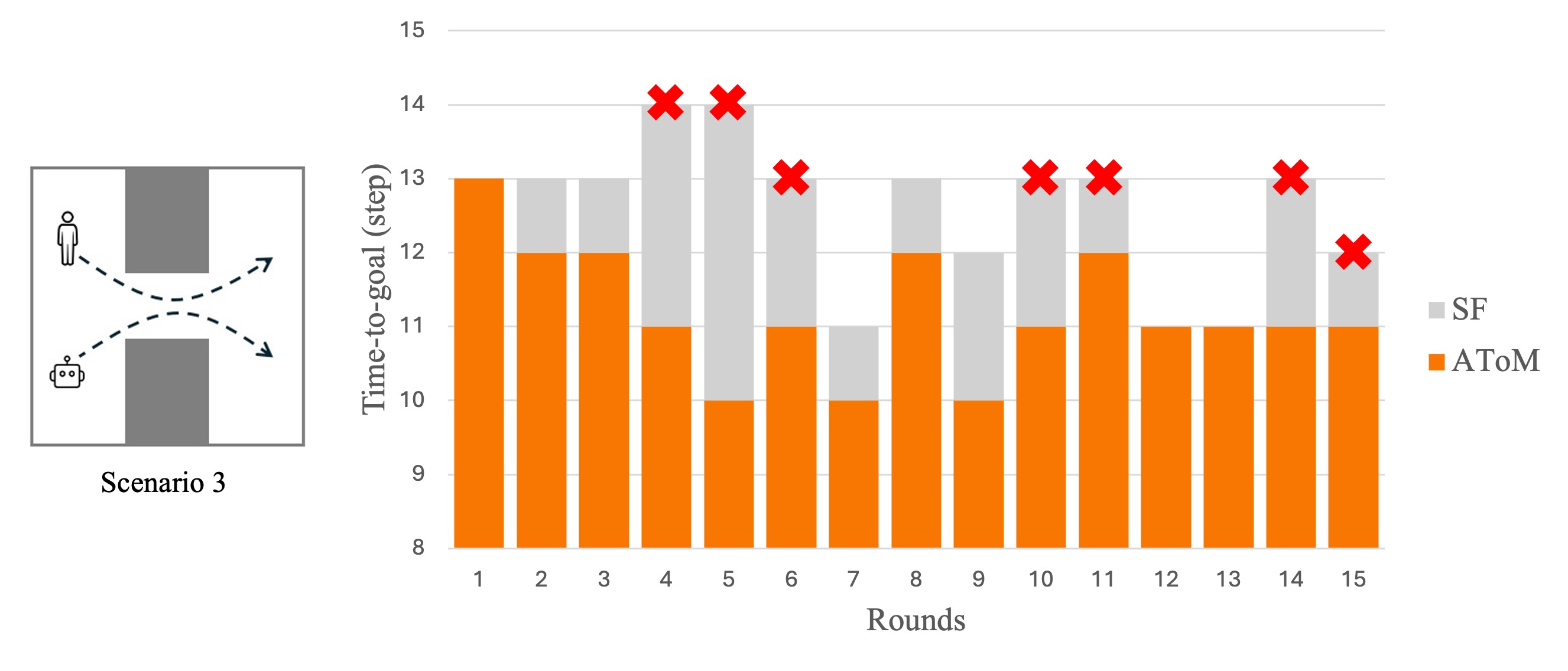}
    \vspace{-8mm}
    \caption{Quantitative comparison between AToM and SF in Scenario 3. We compare the number of steps the robot takes to reach the goal which reflects the efficiency. SF leads to collisions in 7 rounds, which we highlighted using red crosses.}
    % We mark the rounds where collisions happen using red crosses. AToM enables the robot to pass through the doorway efficiently and safely in all rounds of experiments.}
    \vspace{-5mm}
    \label{fig:quantitative_3}
\end{figure}

We then find a Nash equilibrium solution $\tilde{\boldsymbol{U}}_{1}, \dots, \tilde{\boldsymbol{U}}_{n}$ for $n$ players with each player's cost defined in Problem \eqref{eq:game_cost}.
At a Nash equilibrium point, no player $i$ can further decrease its cost by unilaterally changing its strategy $\tilde{\boldsymbol{U}}_{i}$ \cite{facchinei2010generalized}. 
Solving this generalized Nash equilibrium problem can be done using existing dynamic game solvers. In this work, we use ILQSolver \cite{fridovich2020efficient} which iteratively solves linear-quadratic approximations of the original differential game in multi-player settings. 
We consider it an algorithmic module $\GameSolver$ parameterized by $\boldsymbol{\theta}$ that takes as input the current joint state of the system, the known goal positions and environmental obstacles, and returns an open-loop joint strategy that satisfies global Nash equilibrium:
\begin{equation}
\label{eq:our_ToM}
    \tilde{\boldsymbol{U}}_{H}, \tilde{\boldsymbol{U}}_{R} = \GameSolver(\boldsymbol{x}_{HR}^{t}, \boldsymbol{g}_{HR}, \mathcal{O}, \hat{\boldsymbol{\theta}}),
\end{equation}
where $\tilde{\boldsymbol{U}}_{H}$ is the predicted human action and equivalent to $\boldsymbol{\hat{U}}_{H}$, 
$\tilde{\boldsymbol{U}}_{R}$ is the human-predicted robot action, 
and $\hat{\boldsymbol{\theta}}$ is the human-predicted behavioural parameters. 
With this reformulation, we can now easily obtain $\hat{\boldsymbol{X}}_{H}$ from Eq. \eqref{eq:our_ToM} and human dynamics $\Dynamics_{H}$, and therefore solve Problem \eqref{eq:robot_planning} without any recursive structure.
% With this reformulation, we can now substitute Eq. \eqref{eq:our_ToM} and human dynamics back to the robot predictive planning problem in Eq. \eqref{eq:robot_planning} without any recursive structure.

Existing game-theoretic robot planners use similar formulations to solve for the optimal robot plans using known cost functions for each agent \cite{tian2022safety, le2021lucidgames, hu2023emergent}. 
A fundamental difference in this work is that our human internal model $\GameSolver$ is parameterized by $\hat{\boldsymbol{\theta}}$, which represents a dynamic human belief instead of the true agent parameters.

After obtaining predicted human trajectories $\hat{\boldsymbol{X}}_{H}$, the downstream robot planner can now perform predictive planning and execute the actual robot plans. 
As we do not assume any leader-follower structure, the human simultaneously performs her/his actions. 
Up to this point, we have obtained four sets of trajectories: 
1) Predicted Human $\hat{\boldsymbol{X}}_{H}$, 
2) Human-predicted Robot $\hat{\boldsymbol{X}}_{R}$, which can be obtained from $\tilde{\boldsymbol{U}}_{R}$
3) Observed Human $\boldsymbol{X}_{H}$, 
4) Observed Robot $\boldsymbol{X}_{R}$. 
These predicted-observed pairs can be used to correct the estimated behavioural parameters $\hat{\boldsymbol{\theta}}$ in the human internal model.

We use Unscented Kalman Filter (UKF) \cite{wan2000unscented} as the update mechanism. We allow $\hat{\boldsymbol{\theta}}$ to evolve using a random walk as the process model $\ProcessModel$. The measurement states are the agents' trajectories and the measurement model $\MeasurementModel$ is therefore the game solver $\GameSolver$ combined with agent dynamics.
\begin{equation}
\begin{aligned}
    \boldsymbol{\theta}^{t+1} &= \ProcessModel(\boldsymbol{\theta}^{t}, \delta^{t}) = \boldsymbol{\theta}^{t} + \delta^{t}, \quad \delta^t \sim \mathcal{N}(0, Q_t), \\
    \boldsymbol{x}^{t+1} &= \MeasurementModel(\boldsymbol{x}^{t}, \boldsymbol{\theta}^{t}) \\
            &= \Dynamics(\boldsymbol{x}^{t}, \GameSolver(\boldsymbol{x}^{t}, \boldsymbol{\theta}^{t})) + \epsilon^{t}, \quad \epsilon^t \sim \mathcal{N}(0, R),
\end{aligned}
\end{equation}
where $Q_{t}$ and $R$ are the covariance for the process model and measurement model.
In this way, our adaptive ToM human model can be adjusted dynamically to improve its prediction accuracy, and to reflect how humans update their beliefs on others. The complete procedure is detailed in Algorithm \ref{alg:our_method}, where $\Sigma$ is the covariance of $\theta$ estimation.
At the observation step, the robot executes the planned action from the predictive planner and the true human motion is observed.

% \vspace{-1mm}
\begin{algorithm}
\caption{Predict-Observe-Update Procedures with AToM}
\label{alg:our_method}
\begin{algorithmic}[1]
\State \textbf{Inputs:} $\boldsymbol{x}_{HR}^{t}, \hat{\boldsymbol{\theta}}^{t}, \Sigma^{t}, \boldsymbol{g}_{HR}, \mathcal{O}$
\State \textbf{for} $k = t, t+1, \dots$ \textbf{do}
    \State \quad $\tilde{\boldsymbol{U}}_{HR} \gets \GameSolver(\boldsymbol{x}_{HR}^{k}, \boldsymbol{g}_{HR}, \mathcal{O}, \hat{\boldsymbol{\theta}}^{k})$ 
    \State \quad $\hat{\boldsymbol{X}}_{HR} \gets \Dynamics_{HR}(\boldsymbol{x}_{HR}^{k}, \tilde{\boldsymbol{U}}_{HR})$\Comment{Predict}
    \State \quad $\boldsymbol{x}_{R}^{k+1} \gets \text{RobotPlanner}(\boldsymbol{x}_{R}^{k}, \hat{\boldsymbol{X}}_{H}, \boldsymbol{g}_{R}, \mathcal{O})$
    % \State \quad $\boldsymbol{x}_{H}^{k+1} \gets \text{Human}$ \Comment{Observation}
    \State \quad observe human state $\boldsymbol{x}_{H}^{k+1}$ \Comment{Observe}
    \State \quad $\hat{\boldsymbol{\theta}}^{k+1}, \Sigma^{k+1} \gets $
    \Statex \quad $\text{UKF}(\boldsymbol{x}_{HR}^{k+1}, \hat{\boldsymbol{x}}_{HR}^{k+1}, \hat{\boldsymbol{\theta}}^{k}, \Sigma^{k}, \ProcessModel, \MeasurementModel)$ \Comment{Update}
\State \textbf{end for}
\end{algorithmic}
\end{algorithm}
% \vspace{-2mm}

% \begin{equation}
% \begin{aligned}
%     \theta^{t+1} &= \ProcessModel(\theta^{t}, \delta^{t}) \\
%                  &= \theta^{t} + \delta^{t}, \quad \delta^t \sim \mathcal{N}(0, Q_t), \\
%     X^{t+1} &= \MeasurementModel(X^{t}, \theta^{t}) \\
%             &= X^{t} + F(X^{t}, GameSolver_{\theta^{t}}(X^{t})).
% \end{aligned}
% \end{equation}

\section{Experiments}

\subsection{Simulation Settings}
\label{sec:exp_settings}
\textbf{Scenarios.}
We first conduct experiments where the simulated human motions replicate hypothesised human behaviours in long-term repeated interactions, as described below:
% We first evaluate the prediction performance of AToM in three different simulation scenarios. 
% The simulated human motions replicate real-life human behaviours in long-term repeated interactions, as described below.
\begin{enumerate}
    \item \textbf{Scenario 1: 2-agent position exchange.} The robot exchanges positions with the human in a free navigation setting. Across 8 repeated rounds, the simulated human demonstrates an increasing speed while reducing social distance.
    \item \textbf{Scenario 2: 3-agent corridor overtake.} The robot navigates in a narrow corridor with one human in front and another human coming from the opposite direction. Across 8 repeated rounds, we hypothesize that the simulated human in front is pushing a trolley with increasing weights. Therefore they demonstrate a decreasing speed. Their detour is first increased and then decreased. The other simulated human demonstrates an increasing speed while reducing social distance.
    \item \textbf{Scenario 3: 2-agent doorway negotiation.} The robot and the human both need to move through a narrow doorway. Across 15 repeated rounds, the simulated human demonstrates three stages of behaviours, shifting from being ``conservative" to ``cooperative" and to ``aggressive".
\end{enumerate}

% \begin{figure*}[ht]
%     \centering
%     \includegraphics[width=\textwidth]{quantitative_1_2.jpg}
%     \caption{Quantitative comparisons for Scenario 1 and Scenario 2. The simulation setup is illustrated on the left. We compare the prediction accuracy using ADE, and the efficiency and safety in resulting robot plans using Detour and Minimum Distance.}
%     % AToM achieves the best accuracy and results in improving efficiency while maintaining safety.}
%     \vspace{-5mm}
%     \label{fig:quantitative_12}
% \end{figure*}

\begin{figure}
    \centering
    \includegraphics[width=0.8\linewidth]{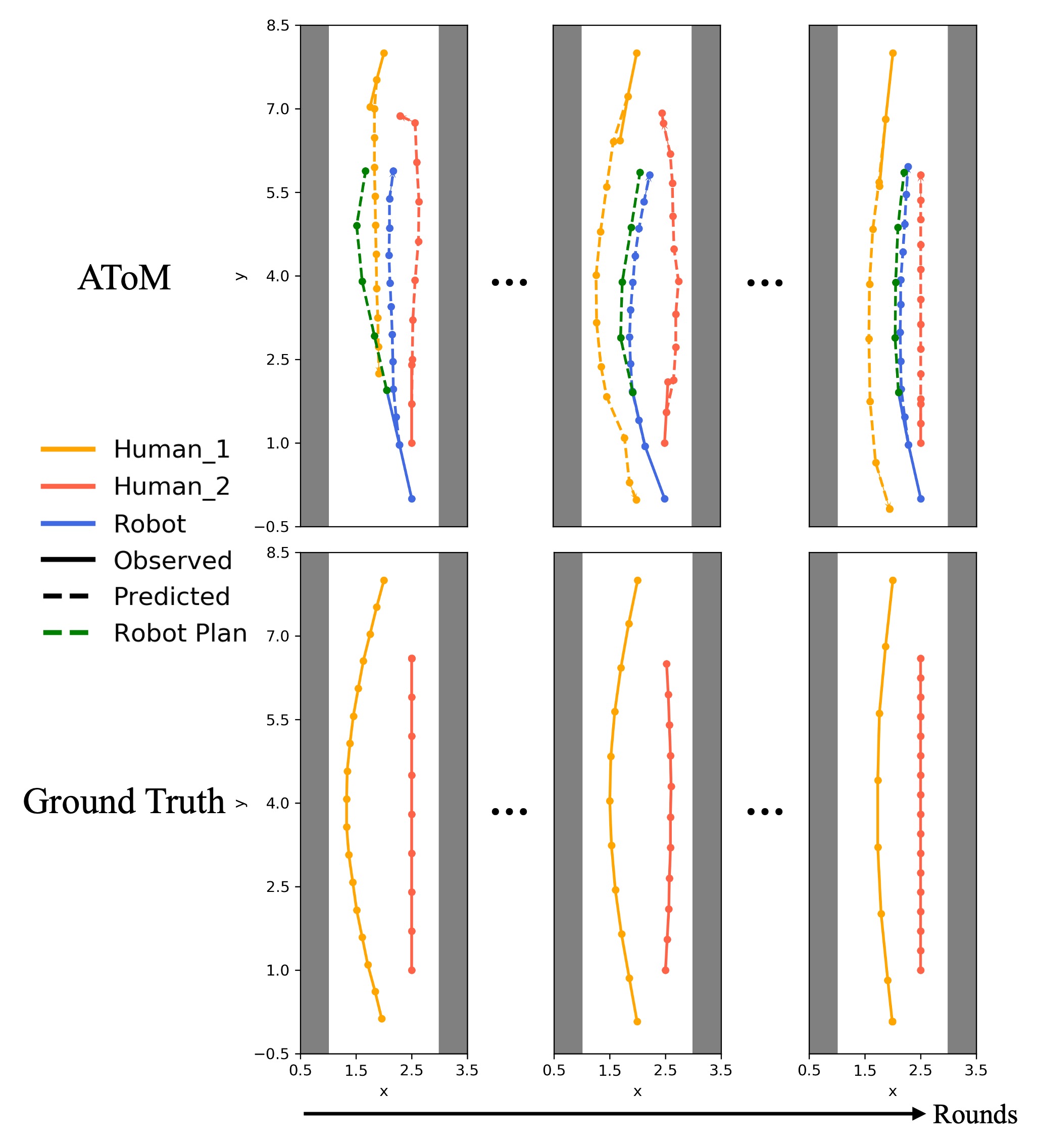}
    \vspace{-3mm}
    \caption{Comparison between the predicted and ground truth trajectories in the first, the middle, and the last round of Scenario 2. Predictions are plotted at the second timestep.}
    % Human behaviours vary between different rounds. AToM is able to capture this change and adjust its predictions to reflect dynamic human internal belief.}
    \vspace{-5mm}
    \label{fig:qualitative_2}
\end{figure}

\textbf{Dynamics.}
Our method is dynamics-agnostic. For ease of implementation, both the human and the robot are modelled using a 2D single integrator where $\dot{x} = x + v\Delta t$. The robot speed is capped at $1.0m/s$ in the first two scenarios and $0.6m/s$ in Scenario 3. The simulated human speed ranges from $0.35m/s$ to $1.2m/s$. For the behavioural parameters, we choose $\boldsymbol{\theta} = [v\_max, d]^\top$ for each agent which controls the maximum speed and preferred social radius. We set process and measurement noise with empirical values.

\textbf{Baselines.}
We compare AToM to four human prediction baselines that are commonly used in social robot navigation:
\begin{enumerate}
    \item Constant Velocity (\textbf{CV}) assumes that humans tend to maintain their speed and direction.
    \item Social Force model (\textbf{SF}) \cite{helbing1995social} calculates human moving direction by a weighted combination of goal attraction, obstacle repulsion, and neighbours repulsion.
    \item \textbf{Trajectron++} \cite{salzmann2020trajectron++} is a dynamics-integrated neural network that considers interactions between pedestrians within a neighboorhood.
    \item \textbf{Memonet} \cite{xu2022remember} is a goal-conditioned neural network that predicts human trajectories based on goal positions.
\end{enumerate}
For SF, we tune the force weights for each scenario. For both Trajectron++ and Memonet, we use the pretrained weight provided by the authors. For Memonet, we skip the goal prediction step and feed the true human goal positions to the trajectory generation sub-network.

\textbf{Implementation Details.}
Our method is planner-agnostic. We choose a recent sampling-based Model Predictive Controller Pred2Nav \cite{poddar2023crowd} as the planning module in all experiments. We execute the first planned action and re-plan at every step. 
% For the first two scenarios, the human trajectories are recorded and replayed when testing with different methods. For the third scenario, the simulated human is keyboard-controlled and follows specific rules for each of the three behaviours: 1) conservative human always allows the robot to pass first, 2) cooperative human only allows the robot to pass first if they moves slower than the robot, 3) aggressive human always passes first. The human speed between each repeated round is different, but this speed variation is the same when testing with different methods. 
All experiments are conducted on a Morefine M9 Pro Mini PC.

\textbf{Evaluation Metrics.} For the first two scenarios, 1) Average Displacement Error (\textbf{ADE}) measures the distance between the true and predicted human trajectories, averaged over all steps, 2) \textbf{Detour} measures how far the robot deviates from the shortest path to goal, averaged over all steps, 3) \textbf{Minimum Distance} measures the closest distance between human and robot within a round. 
% We exclude the prediction at initial timestep for all evaluations because CV requires at least 2 observations to predict. 
For Scenario 3, 
% the robot needs to decide whether to pass the doorway before or after the human. Inaccurate predictions could either confuse the robot to wait for the human unnecessarily or lead to collisions. Therefore, 
we use \textbf{Time-to-goal} to measure how long the robot takes to reach the goal. We consider it a collision if the distance between human and robot is smaller than 0.5m.

% \begin{figure}
%     \centering
%     \includegraphics[width=\linewidth]{quantitative_3.jpg}
%     \caption{Quantitative comparison between AToM and SF in Scenario 3. We compare the number of steps the robot takes to reach the goal which reflects the efficiency. We mark the rounds where collisions happens using red crosses.}
%     % AToM enables the robot to pass through the doorway safely and efficiently in all rounds of experiments.}
%     \vspace{-5mm}
%     \label{fig:quantitative_3}
% \end{figure}

% \begin{figure}
%     \centering
%     \includegraphics[width=0.8\linewidth]{qualitative_2.jpg}
%     \caption{Comparison between the predicted and ground truth trajectories in the first, the middle, and the last round of Scenario 2. Predictions are plotted at the second timestep.}
%     % Human behaviours vary between different rounds. AToM is able to capture this change and adjust its predictions to reflect dynamic human internal belief.}
%     \vspace{-5mm}
%     \label{fig:qualitative_2}
% \end{figure}

\begin{figure*}[ht]
    \centering
    \includegraphics[width=\textwidth]{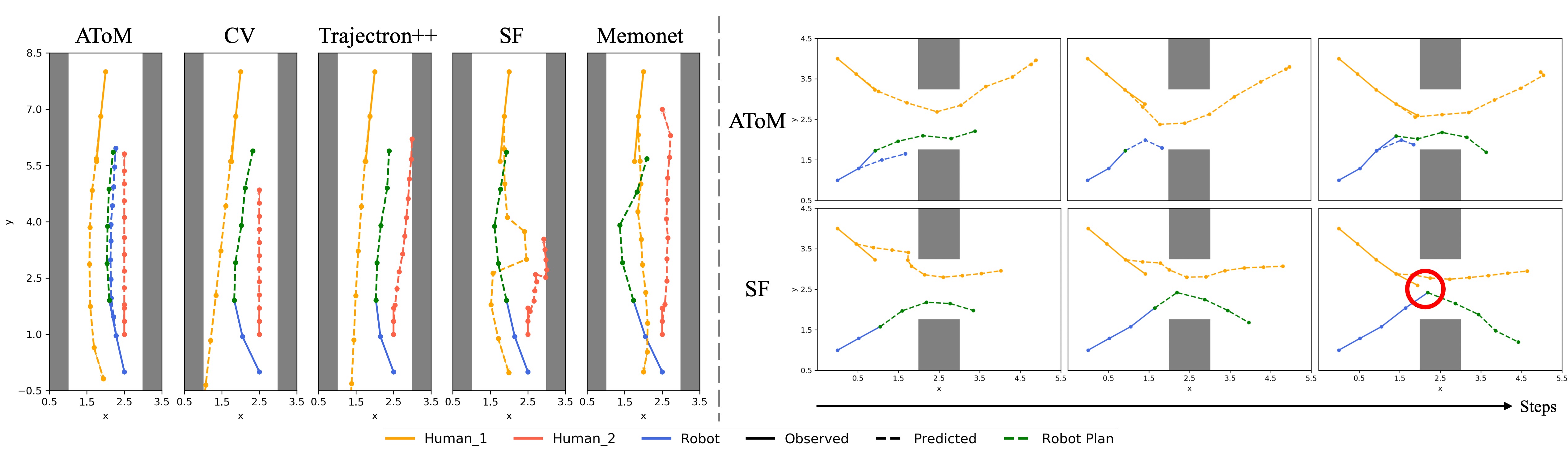}
    \vspace{-5mm}
    \caption{On the left, we compare the predicted trajectories in a sample round from Scenario 2. On the right, we plot three consecutive steps from Scenario 3, where SF prediction misleads the robot into a collision highlighted by the red circle.}
    % \vspace{-2mm}
    \label{fig:failures}
\end{figure*}

\subsection{Quantitative Evaluation}
% The quantitative comparison results are presented in Fig. \ref{fig:quantitative_12} for the first two scenarios, and in Fig. \ref{fig:quantitative_3} for Scenario 3. 
\textbf{Accuracy.}
From Fig. \ref{fig:quantitative_12}-a1 and Fig. \ref{fig:quantitative_12}-b1, we can see that the prediction error of AToM is significantly lower than the baseline methods. 
Memonet shows the worst accuracy in both scenarios as it tends to predict equally spaced waypoints between the current and goal positions which is unrealistic.
CV and Trajectron++ perform similarly and cannot handle situations when the human deviates from a straight path to give way to the robot. CV only shows good accuracy for Human 2 in Scenario 2 whose path is mostly straight. 
SF generates trajectories with unnatural sharp turns when avoiding obstacles and other agents. 
In comparison, AToM models humans that ``look into the future" and can therefore predict smooth and realistic trajectories. 
An important observation is that AToM is the only method that demonstrates decreasing error over the rounds, which shows that our adaptive model becomes more accurate with each update. The decreasing error for human-predicted robot trajectories reflects how humans improve their internal beliefs about the robot over repeated interactions.

\textbf{Efficiency and Safety.}
We now evaluate how human prediction accuracy affects downstream robot planning. We compare efficiency and safety jointly because satisfying one may sacrifice another. 
From Fig. \ref{fig:quantitative_12}-a2-a3 and Fig. \ref{fig:quantitative_12}-b2-b3, the planned robot trajectory using AToM shows decreasing detour which represents increasing efficiency, while maintaining a safe distance from humans. 
Memonet prediction leads to an over-conservative robot, as it models aggressive humans who move straight towards the goal. 
% It fails to maintain a safe distance in later rounds from Scene 2 because the robot falls into a false belief that there is more space on the left side of the corridor, which can be seen later in the qualitative analysis. 
CV and Trajectron++ predictions lead to more efficient robot plans compared to ours, but with a compromise on safety as the minimum distance from human is significantly lower. 
SF shows a similar detour to ours, but its inaccurate prediction leads to untimely avoidance in later rounds when humans are moving fast. 
The effect of inaccurate predictions becomes more noticeable in Scenario 3. We select SF to compare with because obstacle avoidance is critical in this scenario. As shown in Fig. \ref{fig:quantitative_3}, the robot can reach its goal faster with AToM predictions. With inaccurate predictions from SF, the robot tends to wait at the doorway even when the human intends to wait for the robot. 
In 7 out of the 15 rounds when the human is moving with high speed or intends to pass the doorway first, SF model fails to predict the correct human motion and leads to collisions. 
In comparison, our model predicts human behaviours that align with the simulation settings (Sec. \ref{sec:exp_settings}), allowing the robot to wait for the human when necessary to ensure safety.
% In comparison, our model learns that the human is being conservative in the first 5 rounds, gradually allowing the robot to pass without waiting. In the rest of the rounds, our model predicts human intentions accurately, allowing the robot to wait for the human when necessary to ensure safety.

% \begin{figure*}[ht]
%     \centering
%     \includegraphics[width=\textwidth]{failures.jpg}
%     \caption{On the left, we compare the predicted trajectories in a sample round from Scenario 2. On the right, we plot three consecutive steps from Scenario 3, where SF prediction misleads the robot into a collision highlighted by the red circle.}
%     \vspace{-5mm}
%     \label{fig:failures}
% \end{figure*}

\begin{figure}
    \centering
    \includegraphics[width=\linewidth]{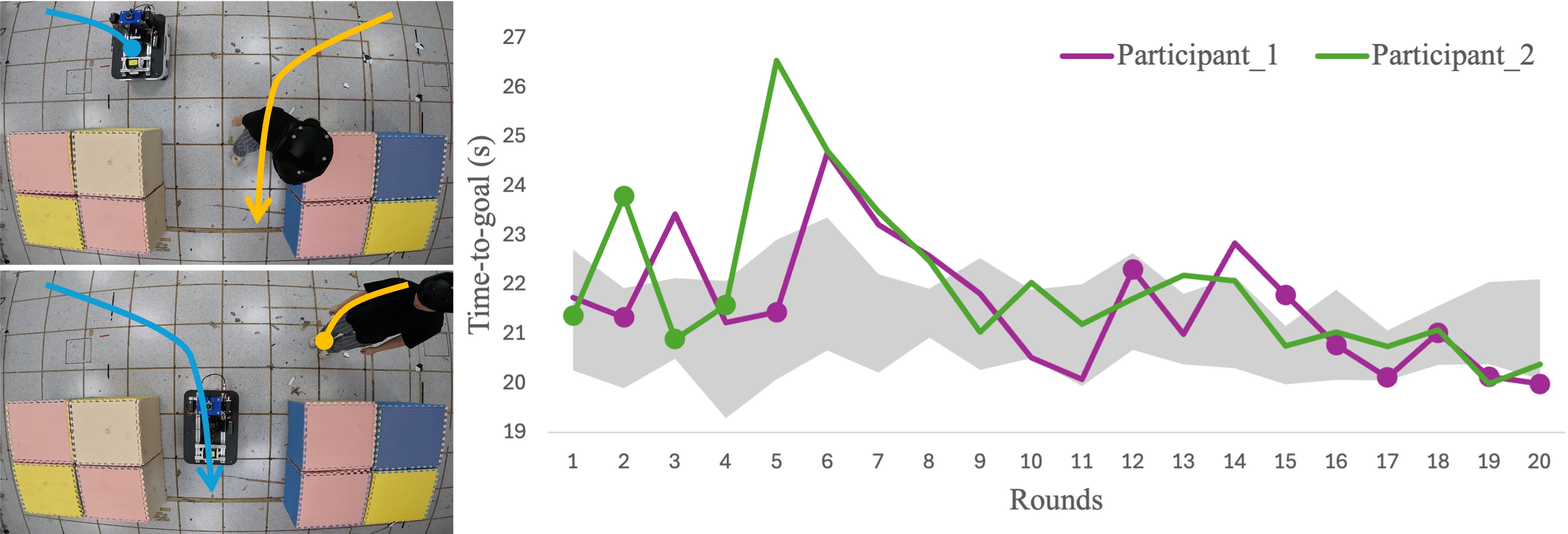}
    \vspace{-5mm}
    \caption{On the left, we demonstrate the 2 situations where either the human or the robot passes the doorway first. On the right, we present the time taken for the robot to reach the goal. We highlight 2 participants who show demonstrative behaviours. The rounds where the human gives way to the robot are labelled with thick dots. The faded areas correspond to the 95\% confidence interval from all participants.}
    \vspace{-3mm}
    \label{fig:hardware}
\end{figure}

\subsection{Qualitative Evaluation}
Fig. \ref{fig:qualitative_2} illustrates predicted trajectories from AToM in 3 different rounds. Our method accurately captures the evolving human behaviours over repeated interactions. 
The initial prediction is inaccurate as it does not capture how Human 1 gives way to the robot. As the experiment continues, the human models are correctly updated to reflect socially aware behaviours and evolving speeds. 
% From round 5 to round 8, Human 1 moves with decreasing detour and increasing speed while human 2 moves back to a straight line with decreasing speed, as we described in \ref{sec:exp_settings}. This change is successfully captured by our dynamic model. 
Since the human-predicted robot trajectory is also getting more accurate, we can say that Human 1 chooses to move more efficiently because they develop a more accurate belief of the robot's collision-avoidance behaviour. This interpretability to dynamic human motions is unique to our AToM model.

Unsatisfactory robot plans using baseline predictions are shown in Fig. \ref{fig:failures}. CV and Trajectron++ overestimate the amount of space and cause the robot to move too close to the actual humans. 
On the contrary, Memonet predicts aggressive humans which forces the robot to take a large detour to avoid both humans. 
SF generates unnatural trajectories that are misleading for the robot. 
% models humans that only react without prediction, resulting in trajectories that are not only unnatural and inaccurate but also misleading for the robot. 
As shown in the example from Scenario 3, SF predicts that the human moves straight towards the wall and makes a sharp turn near the wall boundary. With this false prediction, the robot decides to pass first and a collision occurs. 
In comparison, our method correctly infers that the human intends to pass the doorway first, forcing the robot to wait for the human. From the human's perspective, they predict the robot will wait at the doorway, which explains why the human chooses to pass first with confidence.

\subsection{User Study}
As AToM demonstrates promising performance in simulations, we therefore conduct a real user study for the 2-agent doorway negotiation scenario. 
We recruited 10 participants from the campus community with no prior experience interacting with the robot. Each participant is only given the instruction to walk through the doorway and the experiment is repeated for 20 rounds.
We use Agilex Ranger Mini as the robot platform and Vicon motion capture system to obtain the positions and velocities of each agent.
Sample rounds can be found in the supplementary video.

Different from the pre-defined humans in simulations, real humans each show their unique preferences and evolving behaviours. 
In Fig. \ref{fig:hardware}, the two example participants demonstrate completely different behaviours. What is common is that when the participant shows consistent behaviours (Participant 1 always crosses first after round 5, Participant 2 crosses first during rounds 6-11 but decides to give way after round 15), the robot moves with increasing efficiency. Because AToM predictions get more accurate during these rounds, the robot can decide on the correct move with less confusion. In terms of safety, all rounds are completed without collision.

\section{CONCLUSIONS}

We have presented a novel reformulation of the classic ToM model on human motion prediction. By combining a game-theoretic model with dynamic parameter update, we build a human model that can predict and correct its beliefs about surrounding agents. Our prediction model integrates seamlessly with downstream robot planners and we demonstrate through simulations and real-world experiments that our model not only produces accurate predictions, but also results in safe and efficient robot plans. We rethink human modelling with a fundamental cognitive ability and AToM has provided new insights into long-term human-robot interactions. Future work aims to extend the behavioural parameters to model more complex interaction scenarios.


\bibliographystyle{IEEEtran}
\bibliography{references}

\begin{thebibliography}{10}
\providecommand{\url}[1]{#1}
\csname url@samestyle\endcsname
\providecommand{\newblock}{\relax}
\providecommand{\bibinfo}[2]{#2}
\providecommand{\BIBentrySTDinterwordspacing}{\spaceskip=0pt\relax}
\providecommand{\BIBentryALTinterwordstretchfactor}{4}
\providecommand{\BIBentryALTinterwordspacing}{\spaceskip=\fontdimen2\font plus
\BIBentryALTinterwordstretchfactor\fontdimen3\font minus \fontdimen4\font\relax}
\providecommand{\BIBforeignlanguage}[2]{{%
\expandafter\ifx\csname l@#1\endcsname\relax
\typeout{** WARNING: IEEEtran.bst: No hyphenation pattern has been}%
\typeout{** loaded for the language `#1'. Using the pattern for}%
\typeout{** the default language instead.}%
\else
\language=\csname l@#1\endcsname
\fi
#2}}
\providecommand{\BIBdecl}{\relax}
\BIBdecl

\bibitem{sagheb2023towards}
S.~Sagheb, Y.-J. Mun, N.~Ahmadian, B.~A. Christie, A.~Bajcsy, K.~Driggs-Campbell, and D.~P. Losey, ``Towards robots that influence humans over long-term interaction,'' in \emph{2023 IEEE International Conference on Robotics and Automation (ICRA)}.\hskip 1em plus 0.5em minus 0.4em\relax IEEE, 2023, pp. 7490--7496.

\bibitem{rudenko2020human}
A.~Rudenko, L.~Palmieri, M.~Herman, K.~M. Kitani, D.~M. Gavrila, and K.~O. Arras, ``Human motion trajectory prediction: A survey,'' \emph{The International Journal of Robotics Research}, vol.~39, no.~8, pp. 895--935, 2020.

\bibitem{korbmacher2022review}
R.~Korbmacher and A.~Tordeux, ``Review of pedestrian trajectory prediction methods: Comparing deep learning and knowledge-based approaches,'' \emph{IEEE Transactions on Intelligent Transportation Systems}, vol.~23, no.~12, pp. 24\,126--24\,144, 2022.

\bibitem{sighencea2021review}
B.~I. Sighencea, R.~I. Stanciu, and C.~D. C{\u{a}}leanu, ``A review of deep learning-based methods for pedestrian trajectory prediction,'' \emph{Sensors}, vol.~21, no.~22, p. 7543, 2021.

\bibitem{de2013much}
H.~De~Weerd, R.~Verbrugge, and B.~Verheij, ``How much does it help to know what she knows you know? an agent-based simulation study,'' \emph{Artificial Intelligence}, vol. 199, pp. 67--92, 2013.

\bibitem{de2015higher}
------, ``Higher-order theory of mind in the tacit communication game,'' \emph{Biologically Inspired Cognitive Architectures}, vol.~11, pp. 10--21, 2015.

\bibitem{von2017minds}
F.~B. Von Der~Osten, M.~Kirley, and T.~Miller, ``The minds of many: Opponent modeling in a stochastic game.'' in \emph{IJCAI}, 2017, pp. 3845--3851.

\bibitem{baker2014modeling}
C.~L. Baker and J.~B. Tenenbaum, ``Modeling human plan recognition using bayesian theory of mind,'' \emph{Plan, Activity, and Intent Recognition: Theory and Practice}, vol.~7, no. 177-204, p.~86, 2014.

\bibitem{rabinowitz2018machine}
N.~Rabinowitz, F.~Perbet, F.~Song, C.~Zhang, S.~A. Eslami, and M.~Botvinick, ``Machine theory of mind,'' in \emph{International Conference on Machine Learning}.\hskip 1em plus 0.5em minus 0.4em\relax PMLR, 2018, pp. 4218--4227.

\bibitem{tian2021learning}
R.~Tian, M.~Tomizuka, and L.~Sun, ``Learning human rewards by inferring their latent intelligence levels in multi-agent games: A theory-of-mind approach with application to driving data,'' in \emph{2021 IEEE/RSJ International Conference on Intelligent Robots and Systems (IROS)}.\hskip 1em plus 0.5em minus 0.4em\relax IEEE, 2021, pp. 4560--4567.

\bibitem{cuzzolin2020knowing}
F.~Cuzzolin, A.~Morelli, B.~Cirstea, and B.~J. Sahakian, ``Knowing me, knowing you: theory of mind in ai,'' \emph{Psychological Medicine}, vol.~50, no.~7, pp. 1057--1061, 2020.

\bibitem{schaefer2021leveraging}
S.~Schaefer, K.~Leung, B.~Ivanovic, and M.~Pavone, ``Leveraging neural network gradients within trajectory optimization for proactive human-robot interactions,'' in \emph{2021 IEEE International Conference on Robotics and Automation (ICRA)}.\hskip 1em plus 0.5em minus 0.4em\relax IEEE, 2021, pp. 9673--9679.

\bibitem{tian2022safety}
R.~Tian, L.~Sun, A.~Bajcsy, M.~Tomizuka, and A.~D. Dragan, ``Safety assurances for human-robot interaction via confidence-aware game-theoretic human models,'' in \emph{2022 International Conference on Robotics and Automation (ICRA)}.\hskip 1em plus 0.5em minus 0.4em\relax IEEE, 2022, pp. 11\,229--11\,235.

\bibitem{geldenbott2024legible}
J.~Geldenbott and K.~Leung, ``Legible and proactive robot planning for prosocial human-robot interactions,'' \emph{arXiv preprint arXiv:2404.03734}, 2024.

\bibitem{sripathy2021dynamically}
A.~Sripathy, A.~Bobu, D.~S. Brown, and A.~D. Dragan, ``Dynamically switching human prediction models for efficient planning,'' in \emph{2021 IEEE International Conference on Robotics and Automation (ICRA)}.\hskip 1em plus 0.5em minus 0.4em\relax IEEE, 2021, pp. 3495--3501.

\bibitem{tian2023towards}
R.~Tian, M.~Tomizuka, A.~D. Dragan, and A.~Bajcsy, ``Towards modeling and influencing the dynamics of human learning,'' in \emph{Proceedings of the 2023 ACM/IEEE International Conference on Human-Robot Interaction}, 2023, pp. 350--358.

\bibitem{parekh2023learning}
S.~Parekh and D.~P. Losey, ``Learning latent representations to co-adapt to humans,'' \emph{Autonomous Robots}, vol.~47, no.~6, pp. 771--796, 2023.

\bibitem{muktadir2024adaptive}
G.~M. Muktadir and J.~Whitehead, ``Adaptive pedestrian agent modeling for scenario-based testing of autonomous vehicles through behavior retargeting,'' in \emph{IEEE Int. Conf. Robot. Automat.(ICRA)}, 2024.

\bibitem{cathcart2023proactive}
C.~Cathcart, M.~Santos, S.~Park, and N.~E. Leonard, ``Proactive opinion-driven robot navigation around human movers,'' in \emph{2023 IEEE/RSJ International Conference on Intelligent Robots and Systems (IROS)}.\hskip 1em plus 0.5em minus 0.4em\relax IEEE, 2023, pp. 4052--4058.

\bibitem{wu2019depth}
K.~Wu, M.~A. Esfahani, S.~Yuan, and H.~Wang, ``Depth-based obstacle avoidance through deep reinforcement learning,'' in \emph{Proceedings of the 5th International Conference on Mechatronics and Robotics Engineering}, 2019, pp. 102--106.

\bibitem{wu2018learn}
K.~Wu, M.~Abolfazli~Esfahani, S.~Yuan, and H.~Wang, ``Learn to steer through deep reinforcement learning,'' \emph{Sensors}, vol.~18, no.~11, p. 3650, 2018.

\bibitem{wu2021learn}
K.~Wu, H.~Wang, M.~A. Esfahani, and S.~Yuan, ``Learn to navigate autonomously through deep reinforcement learning,'' \emph{IEEE Transactions on Industrial Electronics}, vol.~69, no.~5, pp. 5342--5352, 2021.

\bibitem{wu2020achieving}
------, ``Achieving real-time path planning in unknown environments through deep neural networks,'' \emph{IEEE Transactions on Intelligent Transportation Systems}, vol.~23, no.~3, pp. 2093--2102, 2020.

\bibitem{wu2019bnd}
------, ``Bnd*-ddqn: Learn to steer autonomously through deep reinforcement learning,'' \emph{IEEE Transactions on Cognitive and Developmental Systems}, vol.~13, no.~2, pp. 249--261, 2019.

\bibitem{wu2019tdpp}
K.~Wu, M.~A. Esfahani, S.~Yuan, and H.~Wang, ``Tdpp-net: Achieving three-dimensional path planning via a deep neural network architecture,'' \emph{Neurocomputing}, vol. 357, pp. 151--162, 2019.

\bibitem{cao2022direct}
K.~Cao, M.~Cao, S.~Yuan, and L.~Xie, ``Direct: A differential dynamic programming based framework for trajectory generation,'' \emph{IEEE Robotics and Automation Letters}, vol.~7, no.~2, pp. 2439--2446, 2022.

\bibitem{helbing1995social}
D.~Helbing and P.~Molnar, ``Social force model for pedestrian dynamics,'' \emph{Physical Review E}, vol.~51, no.~5, p. 4282, 1995.

\bibitem{salzmann2020trajectron++}
T.~Salzmann, B.~Ivanovic, P.~Chakravarty, and M.~Pavone, ``Trajectron++: Dynamically-feasible trajectory forecasting with heterogeneous data,'' in \emph{Computer Vision--ECCV 2020: 16th European Conference, Glasgow, UK, August 23--28, 2020, Proceedings, Part XVIII 16}.\hskip 1em plus 0.5em minus 0.4em\relax Springer, 2020, pp. 683--700.

\bibitem{xu2022remember}
C.~Xu, W.~Mao, W.~Zhang, and S.~Chen, ``Remember intentions: Retrospective-memory-based trajectory prediction,'' in \emph{Proceedings of the IEEE/CVF Conference on Computer Vision and Pattern Recognition}, 2022, pp. 6488--6497.

\bibitem{cao2024learningdynamicweightadjustment}
\BIBentryALTinterwordspacing
M.~Cao, X.~Xu, Y.~Yang, J.~Li, T.~Jin, P.~Wang, T.-Y. Hung, G.~Lin, and L.~Xie, ``Learning dynamic weight adjustment for spatial-temporal trajectory planning in crowd navigation,'' 2024. [Online]. Available: \url{https://arxiv.org/abs/2412.00555}
\BIBentrySTDinterwordspacing

\bibitem{boldrer2022multi}
M.~Boldrer, A.~Antonucci, P.~Bevilacqua, L.~Palopoli, and D.~Fontanelli, ``Multi-agent navigation in human-shared environments: A safe and socially-aware approach,'' \emph{Robotics and Autonomous Systems}, vol. 149, p. 103979, 2022.

\bibitem{boldrer2020socially}
M.~Boldrer, M.~Andreetto, S.~Divan, L.~Palopoli, and D.~Fontanelli, ``Socially-aware reactive obstacle avoidance strategy based on limit cycle,'' \emph{IEEE Robotics and Automation Letters}, vol.~5, no.~2, pp. 3251--3258, 2020.

\bibitem{ryu2024integrating}
K.~Ryu and N.~Mehr, ``Integrating predictive motion uncertainties with distributionally robust risk-aware control for safe robot navigation in crowds,'' \emph{arXiv preprint arXiv:2403.05081}, 2024.

\bibitem{poddar2023crowd}
S.~Poddar, C.~Mavrogiannis, and S.~S. Srinivasa, ``From crowd motion prediction to robot navigation in crowds,'' in \emph{2023 IEEE/RSJ International Conference on Intelligent Robots and Systems (IROS)}.\hskip 1em plus 0.5em minus 0.4em\relax IEEE, 2023, pp. 6765--6772.

\bibitem{facchinei2010generalized}
F.~Facchinei and C.~Kanzow, ``Generalized nash equilibrium problems,'' \emph{Annals of Operations Research}, vol. 175, no.~1, pp. 177--211, 2010.

\bibitem{fridovich2020efficient}
D.~Fridovich-Keil, E.~Ratner, L.~Peters, A.~D. Dragan, and C.~J. Tomlin, ``Efficient iterative linear-quadratic approximations for nonlinear multi-player general-sum differential games,'' in \emph{2020 IEEE International Conference on Robotics and Automation (ICRA)}.\hskip 1em plus 0.5em minus 0.4em\relax IEEE, 2020, pp. 1475--1481.

\bibitem{le2021lucidgames}
S.~Le~Cleac’h, M.~Schwager, and Z.~Manchester, ``Lucidgames: Online unscented inverse dynamic games for adaptive trajectory prediction and planning,'' \emph{IEEE Robotics and Automation Letters}, vol.~6, no.~3, pp. 5485--5492, 2021.

\bibitem{hu2023emergent}
H.~Hu, K.~Nakamura, K.-C. Hsu, N.~E. Leonard, and J.~F. Fisac, ``Emergent coordination through game-induced nonlinear opinion dynamics,'' in \emph{2023 62nd IEEE Conference on Decision and Control (CDC)}.\hskip 1em plus 0.5em minus 0.4em\relax IEEE, 2023, pp. 8122--8129.

\bibitem{wan2000unscented}
E.~A. Wan and R.~Van Der~Merwe, ``The unscented kalman filter for nonlinear estimation,'' in \emph{Proceedings of the IEEE 2000 Adaptive Systems for Signal Processing, Communications, and Control Symposium (Cat. No. 00EX373)}.\hskip 1em plus 0.5em minus 0.4em\relax IEEE, 2000, pp. 153--158.

\end{thebibliography}

\end{document}